\documentclass{article}
\usepackage[utf8]{inputenc}
\usepackage[letterpaper,top=2cm,bottom=2cm,left=3cm,right=3cm,marginparwidth=1.75cm]{geometry}
\usepackage{makecell}
\usepackage{graphicx}
\usepackage{subcaption}
\usepackage{amsmath}
\usepackage{pifont}
\usepackage{booktabs}

\title{Preventing Data Leakage in EEG-Based Survival Prediction: A Two-Stage Embedding and Transformer Framework}
\author{
  Yixin Zhou \\
  University of Pittsburgh\\
  School of Computing and Information\\
  Pittsburgh, PA 15213 \\
  \texttt{yiz277@pitt.edu} \\ \and
  Zhixiang Liu \\
  University of Pittsburgh\\
  School of Computing and Information\\
  Pittsburgh, PA 15213 \\
  \texttt{zhl136@pitt.edu} \\ \and
  Vladimir I. Zadorozhny \\
  University of Pittsburgh\\
  School of Computing and Information\\
  Pittsburgh, PA 15213 \\
  \texttt{viz@pitt.edu} \\ \and
    Jonathan Elmer \\
  University of Pittsburgh\\
  Department of Critical Care Medicine \\
  Pittsburgh, PA 15213 \\
  \texttt{elmerjp@upmc.edu}
}
\date{June 2025}

\begin{document}

\maketitle

\begin{abstract}
Deep learning models have shown promise in EEG-based outcome prediction for comatose patients after cardiac arrest, but their reliability is often compromised by subtle forms of data leakage. In particular, when long EEG recordings are segmented into short windows and reused across multiple training stages, models may implicitly encode and propagate label information, leading to overly optimistic validation performance and poor generalization.

In this study, we identify a previously overlooked form of data leakage in multi-stage EEG modeling pipelines. We demonstrate that violating strict patient-level separation can significantly inflate validation metrics while causing substantial degradation on independent test data. 

To address this issue, we propose a leakage-aware two-stage framework. In the first stage, short EEG segments are transformed into embedding representations using a convolutional neural network with an ArcFace objective. In the second stage, a Transformer-based model aggregates these embeddings to produce patient-level predictions, with strict isolation between training cohorts to eliminate leakage pathways.

Experiments on a large-scale EEG dataset of post–cardiac-arrest patients show that the proposed framework achieves stable and generalizable performance under clinically relevant constraints, particularly in maintaining high sensitivity at stringent specificity thresholds. These results highlight the importance of rigorous data partitioning and provide a practical solution for reliable EEG-based outcome prediction.
\end{abstract}

\section{Introduction}

Cardiac arrest remains one of the leading causes of death worldwide \cite{eegInComa2000}. Patients who survive initial resuscitation are often admitted to the intensive care unit (ICU) in a comatose state, where clinicians face the critical challenge of predicting neurological recovery. Distinguishing patients who are likely to regain consciousness from those with irreversible brain injury is inherently difficult, yet such decisions carry significant clinical, ethical, and economic implications.

Premature withdrawal of life-sustaining therapy may lead to preventable deaths, whereas prolonged treatment of non-recoverable patients imposes substantial emotional and financial burdens on families and healthcare systems. Consequently, there is a strong need for reliable decision-support systems that can assist clinicians in making outcome predictions under stringent safety constraints, particularly requiring near-zero false reassurance for patients who may still recover.

Electroencephalography (EEG) has emerged as a key modality for assessing brain function in comatose patients \cite{eegInComa2000}. EEG signals are highly sensitive to the severity and progression of brain dysfunction, especially when continuous or serial recordings are available. Prior studies have demonstrated that EEG measurements obtained within the first 24 hours after cardiac arrest provide substantial prognostic value for both favorable and poor neurological outcomes \cite{Hofmeijer2015}.

However, EEG data are typically high-dimensional and recorded over extended periods using multiple channels, making direct analysis challenging. The scale and complexity of such data introduce significant computational and methodological barriers, motivating the development of structured data infrastructures such as BrainFlux, which aggregates and organizes large-scale EEG data for downstream analysis \cite{BrainFlux2019}.

\paragraph{Contributions.}
In this study, we investigate EEG-based outcome prediction under strict clinical constraints and identify a previously overlooked form of data leakage in multi-stage modeling pipelines. However, direct end-to-end modeling proved infeasible due to the variable length and extended duration of EEG recordings, as well as the high risk of patient-level data leakage—a problem we initially overlooked. Early experiments that ignored this issue reported suspiciously high validation metrics (e.g., AUC $>$ 0.95, Sensitivity@99\% $>$ 0.85), but later collapsed on the independent test set (e.g., Sensitivity@99\% $\approx$ 0.32), revealing severe overfitting. We describe this issue and its mitigation in detail in Section 3.5. 

To address this problem, we propose a two-stage framework. First, neural networks are used to learn embeddings from 5-minute EEG slices. Then, a Transformer-based sequence model is trained on an independently partitioned cohort to capture long-term temporal patterns while preventing data leakage. The proposed method achieves stable and generalizable performance in predicting post–cardiac-arrest outcomes, with improvements over prior approaches \cite{Elmer2019groupbased}.

\section{Related Work}
The analysis work of others suggested that EEG data can be a useful indicator of a patient's health condition, valid interpretation and analysis of EEG should be approved by the medical community for the legal declaration of human death. In France, the law requires that the clinical diagnosis of brain death must be confirmed by two inactive electroencephalograms or cerebral angiograms performed within 4 hours. 
Deep coma and brain death are also very different in EEG. Others applied several statistical complexity measures to the EEG signal and evaluated the differences between two groups of patients: subjects in deep coma and those classified as Brain-dead subjects. In such clinical studies, significant differences between the former and real-life EEG recordings have been reported.\cite{BrainDeathDiagnosis2008}
And others also have done some similar work to prove that it is feasible to use EEG to determine whether people will wake up after a deep coma. Studies have shown that the diversity and variability of EEG of non-survivors are significantly lower than those of survivors; those with predominant delta and slow theta oscillations, either alone or in combination, were found to be more likely to survive during the first evaluation than in patients with a poor outcome (i.e., death) within six months of injury (i.e., death).\cite{LifeorDeat2011}

\section{Materials and Methods}
\subsection{Dataset and Data Collection}

The dataset used in this study consists of EEG signal data exported from the BrainFlux data warehouse. As introduced in the previous section, EEG is an informative data source that reflects the electrical activity of the patient's brain. Each data point is associated with a specific timestamp to maintain temporal information, helping physicians detect abnormalities in patient brain waves and make clinical decisions accordingly.

This dataset contains multiple EEG-derived measurements across different feature types. 

The dataset contains 1,231 patients in total and we split them into three subsets, 581 patients for the training set, 150 patients for the independently partitioned training subset and 500 patients for the test set. The splitting was performed at the patient's level to prevent data leakage. Notably, the ratio of non-survivors to survivors in the dataset is approximately 3:1, indicating a notable class imbalance that must be considered.

\subsection{Task Description}
This study focuses on one particular task: Given electrical measurements of patients' EEG data, predict whether patients will wake up or not after cardiac arrest. The label of past patients in the dataset is given by a feature called "surv", where 1 means this patient woke up in ICU eventually, and 0 means otherwise.
In this particular task, common performance metrics in machine learning including accuracy, Area Under the Curve (AUC), F1 score are not suitable, the optimal model should be extremely good at identifying poor outcomes. Withdrawing treatment based on model predictions is a critical clinical decision, it's unacceptable for all people concerned, especially when the patient could have survived if proper life-sustaining therapy is given. On the other hand, continuing life-sustaining therapy on patients while monitoring their status can be viewed as an acceptable measurement, even if the patients die eventually. Thus, patients with poor outcomes should have more weight than patients otherwise in the dataset, and the final model of this task should have great predictive power on poor outcome cases with near-zero misclassification on patients who recover. Formally, we consider \textbf{Type 2 error} (patients who recover eventually been predicted as poor outcomes) to be much more severe than  \textbf{Type 1 error} ( patients with poor outcomes been predicted as positive outcomes), therefore, a desired model should maximize sensitivity while maintaining high specificity. That said, the particular performance metric for this task is sensitivity when specificity is at a fixed strict threshold (e.g. 95\% or 99\%).

\subsection{Data preprocessing methods}
A standard approach to preprocessing data in time series modeling is Min-Max normalization. This method has been commonly applied in the context of non-stationary signals as discussed by Ogasawara et al. \cite{ogasawara2010adaptive}.

However, during preliminary analysis of our dataset, we observed a significant heavy-tailed distribution on the upper end, which made Min-Max normalization unsuitable for our dataset. While most values were close to the mean, rare but extreme outliers—often hundreds of times larger than the mean value—would disproportionately stretch the scale. This causes most values to be close to zero and leads to a loss of accuracy in the normalized data.

After inspection, we found that these outliers were often caused by measurement artifacts, typically occurring during abrupt disconnection or adjustment of EEG electrodes. Though such noise is clinically irrelevant, it can strongly distort the data range and compromise time series modeling.

To solve this problem, we evaluated a set of normalization strategies that include logarithmic compression and percentile-based truncation. Specifically, we tested: (1) Min-Max normalization using the global minimum and maximum values of the dataset as a baseline, (2) percentile-truncated normalization using the 1st–99th or 5th–95th percentiles, and (3) logarithmic variants of both. Each method was applied and evaluated based on its effect on model performance.

We used a lightweight baseline model for evaluation. The Rhythmicity Spectrogram in the 0–8 Hz frequency band was selected as input because it reflects clinically meaningful EEG activity and has been previously used in seizure detection pipelines \cite{handa2021seizure}, while still containing extreme outliers. The model contains only a simple 1D convolutional layer followed by a dense layer for binary classification. The resulting validation loss and accuracy are summarized in Table~\ref{tab:norm_strategies}. Among all candidates, Experiment 3—Min-Max normalization using the 1st–99th percentiles—achieved the best validation performance and was adopted for all subsequent experiments.

\begin{table}[h]
\centering
\begin{tabular}{|c|c|c|c|c|c|}
\hline
\textbf{Exp. No.} & \textbf{Log} & \textbf{Truncation} & \textbf{Percentile Range} & \textbf{Min Val Loss} & \textbf{Max Val Acc} \\
\hline
1 & \ding{55} & \ding{55} & 0\%--100\% & did not converge & did not converge \\
2 & \ding{51} & \ding{55} & 0\%--100\% & 0.5684 & 0.7158 \\
3 & \ding{55} & \ding{51} & 1\%--99\%  & \textbf{0.5398} & \textbf{0.7569} \\
4 & \ding{51} & \ding{51} & 1\%--99\%  & 0.5964 & 0.7039 \\
5 & \ding{55} & \ding{51} & 5\%--95\%  & 0.5541 & 0.7423 \\
6 & \ding{51} & \ding{51} & 5\%--95\%  & 0.5774 & 0.7167 \\
\hline
\end{tabular}
\caption{Comparison of six normalization strategies combining log transformation and percentile truncation. A checkmark (\ding{51}) indicates the use of a given technique, while a cross (\ding{55}) indicates it was not applied.}
\label{tab:norm_strategies}
\end{table}

\subsection{Feature Engineering}

Based on our previous model, we created several model variants, each sized to accommodate input features of different dimensionalities. This design enables every variant to capture the salient characteristics of its assigned feature group while controlling model capacity and, consequently, the risk of over-fitting.

\begin{figure}[htbp]
  \centering
  \includegraphics[width=0.95\textwidth]{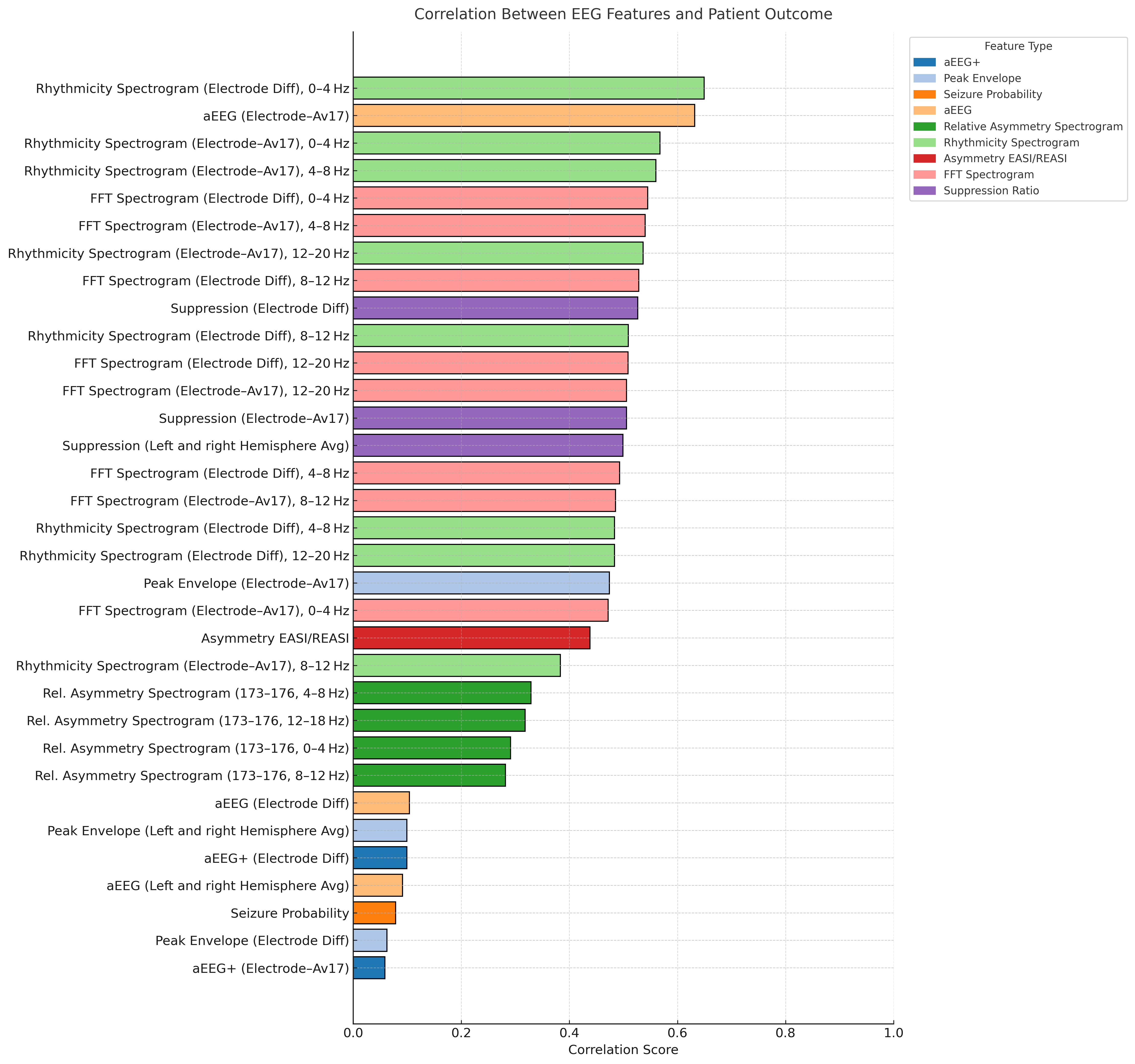}
  \caption{Correlation between EEG features and patient outcomes.}
  \label{fig:eeg_feature_corr}
\end{figure}

Input groups were defined according to canonical EEG frequency bands—\emph{delta} (0.5\,–\,4 Hz), \emph{theta} (4\,–\,8 Hz), etc.—each representing a physiologically meaningful range.  
By combining these bands across the available feature types, we produced 33 distinct input configurations.  
Each variant was trained on the training set with an adaptive learning rate and early stopping, and its best‐performing weights were selected on the validation set. To quantify the association between a given input configuration and patient outcome, we used a normalized correlation score,  
\[
\text{Correlation Score}=2\times\bigl(\text{Balanced Accuracy}-0.5\bigr),
\]
which rescales balanced accuracy to the interval $[0,1]$ and is robust to class imbalance.  
The correlation scores for all evaluated configurations are visualized in Figure~\ref{fig:eeg_feature_corr}.

By ranking the correlation scores, we isolated the EEG feature sets that exhibit the strongest association with patient outcome. The next sections detail which of these features were retained for the final sliding-window model and explain how different feature types were integrated.

We also computed Pearson correlations within each feature family. As expected, features drawn from adjacent frequency bands showed marked intra-group collinearity, reflecting the natural redundancy that arises when measurements differ only by spectral sub-band.

\subsection{Two-Stage Training and Data Leakage Prevention}

Based on the correlation scores presented in Section 2.4, we retained the feature groups with the strongest associations to patient survival. These selected features constitute the input to the subsequent two-stage training pipeline:  
(1) \textbf{FFT Spectrogram} (adjacent-electrode differences, 0–4 Hz),  
(2) \textbf{aEEG} (electrode AV17),  
(3) \textbf{Rhythmicity Spectrogram} (adjacent-electrode differences, 0–4 Hz), and  
(4) \textbf{Suppression Ratio} (adjacent-electrode differences).

\paragraph{Motivation.}
Post–cardiac-arrest EEG traces span many hours and differ markedly in length across patients.  Feeding an entire patient-level signal to a neural network in one pass is not only computationally prohibitive but also statistically unsound.  A naïve workaround is to split each record into overlapping windows and treat those windows as independent samples; however, if windows from the \emph{same} patient are scattered across training, validation, and test folds, models gain an unfair preview of the evaluation data—a classical form of data leakage.  
\textbf{More subtly, if a window that helped optimise an encoder in Stage 1 is \emph{reused} to train a sequence-level model in Stage 2, the second-stage network can ``memorise'' the label information already embedded by the first stage, inflating validation scores while crippling generalisation.}

\paragraph{Stage 1 – Local feature extraction.}
Each EEG record is traversed with a 5-minute sliding window (stride = 5 min).  
Windows are processed by a lightweight 1-D CNN whose output layer is equipped with an \emph{ArcFace} head \cite{deng2019arcface}.  
The additive-angular-margin loss drives class centres apart on a hypersphere and yields compact intra-class clusters.  
The penultimate layer (hereafter the \emph{embedding vector}) is therefore forced to encode discriminative temporal patterns present in short EEG segments.  
Although the Stage-1 network outputs a window-level survival probability, our principal objective in this stage is to harvest the embeddings as fixed-length representations of local brain dynamics.

Stage-1 is trained on 581 patients and validated on an \emph{exclusive} cohort of 150 patients.  No patient overlap is permitted between these two sets.

\paragraph{Stage 2 – Sequence modeling without leakage.}
To eliminate the leakage pathway described above, Stage 2 is trained only on embeddings generated from patients that were never seen during Stage-1 training.
Concretely, the 150-patient validation cohort from Stage 1 is partitioned into an 60/40 split:  
90 patients supply embedding sequences for Stage-2 training, and the remaining 60 patients form a Stage-2 validation set used for early stopping.  
A transformer encoder with 1 layer, 8 attention heads, and positional encoding ingests each patient’s embedding sequence and outputs a survival probability at the patient's level.

Any window that contributed to the optimization of the \emph{ArcFace}-CNN is irrevocably excluded from the Stage-2 training pool.  
In control experiments where this constraint was intentionally violated, validation AUC increased by 0.08, while test-set AUC dropped from 0.909 to 0.730 and Sensitivity@99\% decreased from 0.618 to 0.324. This divergence between validation and test performance highlights the presence of hidden leakage and underscores the necessity of strict patient-level segregation.

\begin{figure}[htbp]
  \centering
  \includegraphics[width=0.95\textwidth]{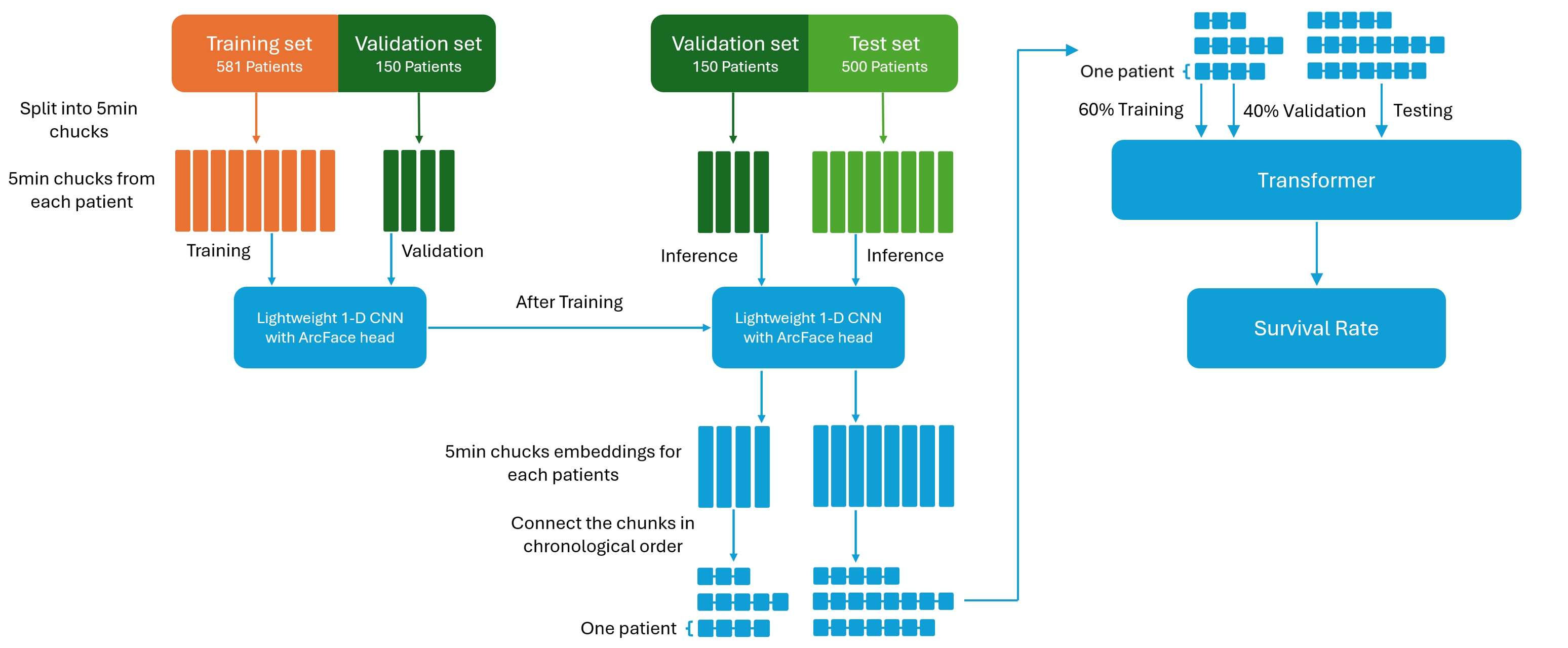}
  \caption{Schematic of the two-stage training framework.}
  \label{fig:twostage_scheme}
\end{figure}

% ---------- Final evaluation ----------
\paragraph{Final evaluation.}
After Stage-2 training converges, both stages are frozen and evaluated on an independent test cohort of 500 patients that was never exposed to either network during optimization.  This hierarchical design (illustrated in Figure \ref{fig:twostage_scheme}) guarantees that no fragment of EEG used to update network weights appears in the model-selection or testing phases, thereby providing a trustworthy estimate of real-world performance.

\paragraph{Control experiment on leakage.}
To quantify the impact of patient-level leakage, we intentionally allowed EEG windows from the \emph{same} patients to appear in both Stage-1 and Stage-2 training folds.  
With this leaky setup, validation metrics looked nearly flawless (AUC $>\!0.99$, Sensitivity@99 \% $>\!0.80$), yet performance collapsed on the independent test cohort (AUC = 0.900; Sensitivity@99 \% = 0.324).  
When the strict patient-level partitioning was reinstated, validation and test results aligned closely (AUC = 0.909; Sensitivity@99 \% = 0.618), confirming that rigorous segregation of patients across stages is essential for trustworthy generalisation.

\subsection{Evaluation Metrics}
Several standard classification metrics were used to evaluate the final result at the patient level. Given the substantial clinical cost of withdrawing life-sustaining therapy from patients who might recover, metrics which emphasize sensitivity under strict constraints on specificity were also added.

The following metrics are reported:

\begin{itemize}
  \item \textbf{AUC (Area Under the ROC Curve):} Reflects the model's ability to discriminate between survivors and non-survivors across all thresholds.

  \item \textbf{Accuracy (ACC):} Proportion of correct predictions across all patients.

  \item \textbf{Precision and Recall:} Precision reflects the proportion of predicted positives that are correct; recall (also known as sensitivity) measures the proportion of true positives that are identified.

  \item \textbf{F1 Score:} Harmonic mean of precision and recall, included to account for class imbalance.

  \item \textbf{Sensitivity@Spec95 and Sensitivity@Spec99:} Measures recall when specificity is constrained to at least 95\% or 99\%, respectively. These metrics reflect the clinical need to minimize false positives when identifying patients with a high likelihood of death.
\end{itemize}

All metrics are computed for three non-overlapping cohorts:

\begin{itemize}
  \item Stage-2 training set (90 patients)
  \item Stage-2 validation set (60 patients)
  \item Independent test set (500 patients)
\end{itemize}

\section{Results}

\paragraph{Leakage-Free Two-Stage Framework Yields Strong Generalization.}

Our two-stage pipeline—local embeddings from 5-minute windows followed by a patient-level Transformer—was trained and evaluated on three non-overlapping cohorts: a 90-patient training cohort, a 60-patient validation cohort, and a 500-patient independent test cohort.  All metrics are computed per patient; 95\% confidence intervals (CI) were estimated by non-parametric bootstrapping over patient IDs (1,000 resamples).

\begin{table}[h]
\centering
\caption{Patient-level performance with 95\% confidence intervals.}
\label{tab:perf_ci}
\begin{tabular}{lccc}
\toprule
\textbf{Metric} & \textbf{Train} & \textbf{Validation} & \textbf{Test} \\
\midrule
Accuracy            & 0.932\,[0.875–0.977] & 0.864\,[0.763–0.949] & 0.844\,[0.812–0.876] \\
AUC                 & 0.980\,[0.948–1.000] & 0.955\,[0.894–1.000] & 0.909\,[0.883–0.933] \\
Precision           & 0.968\,[0.921–1.000] & 0.892\,[0.781–1.000] & 0.884\,[0.849–0.915] \\
Recall              & 0.938\,[0.873–0.985] & 0.892\,[0.786–0.974] & 0.889\,[0.855–0.920] \\
F1 Score            & 0.952\,[0.912–0.985] & 0.892\,[0.810–0.961] & 0.887\,[0.861–0.911] \\
Sensitivity@95\%    & 0.812\,[0.662–1.000] & 0.757\,[0.540–1.000] & 0.711\,[0.617–0.798] \\
Sensitivity@99\%    & 0.734\,[0.641–1.000] & 0.649\,[0.514–1.000] & 0.618\,[0.407–0.687] \\
\bottomrule
\end{tabular}
\end{table}

\paragraph{Independent test-set performance.}
On the 500-patient test cohort the model attained an AUC of \textbf{0.909} (95\% CI: 0.883–0.933) and an F1 score of \textbf{0.887} (0.861–0.911).  
Crucially, at the clinically stringent threshold of 99\% specificity the sensitivity reached \textbf{0.618} (0.407–0.687), and rose to \textbf{0.711} (0.617–0.798) when the specificity requirement was relaxed to 95\%.  
These results demonstrate that the proposed framework can identify a substantial proportion of poor-outcome patients while keeping false-positive rates within clinically acceptable bounds.

\paragraph{Leakage prevention and generalisation.}
The tight overlap between validation- and test-set confidence intervals (e.g., AUC 0.955 vs.\ 0.909; Sensitivity@99\% 0.649 vs.\ 0.618) indicates that the model generalises well and suffers minimal over-fitting.  
This stability confirms that isolating the Stage-2 training cohort from any patient used to optimise Stage 1 successfully prevents label leakage—precisely the problem highlighted in Section 1.2—and yields performance that holds on an unseen, real-world cohort. In contrast, models trained under the leaky setup produced validation AUCs exceeding 0.95 and Sensitivity@99\% nearing 0.90, yet collapsed to 0.900 and 0.324 respectively on the test set. This stark contrast reinforces that naïve window-based splitting creates misleading validation metrics and severely overstates model reliability.

\paragraph{Clinical implication.}
Combined with its lightweight 5-minute window encoder, the two-stage architecture provides a practical decision-support tool: once trained, it requires only forward passes on embeddings and a single Transformer inference to deliver patient-level prognoses with high specificity—meeting the near-zero false-reassurance requirement stated in our introduction.

\section{Discussion and Conclusion}

\paragraph{Key contributions.}
We introduced a leakage-preventive, two-stage framework for predicting neurological recovery in post–cardiac-arrest patients from long-duration EEG.  Stage 1 converts 5-minute windows into compact embeddings; Stage 2 aggregates these embeddings with a lightweight Transformer trained on a strictly disjoint patient cohort.  This design tackles three long-standing obstacles in ICU EEG modelling: ultra-long sequences, severe class imbalance, and—most importantly—patient-level data leakage.

\paragraph{Lessons on data leakage.}
Our early experiments, which unknowingly reused encoder-trained patients in the sequence model, produced deceptively high validation scores (Sensitivity\textsubscript{@99\,\%}~\(\approx0.90\)) yet collapsed on the independent test set (0.32).  This stark gap illustrates how easily subtle leakage can inflate internal metrics and undermine clinical reliability.  By enforcing hard patient boundaries between stages, validation and test performance converged (e.g.\ AUC 0.955 vs 0.909; Sensitivity\textsubscript{@99\,\%} 0.649 vs 0.618), providing a trustworthy estimate of real-world utility.

\paragraph{Clinical relevance.}
Combined with a 5-minute window encoder, the final model delivers patient-level prognoses with 99 \% specificity after a single forward pass—satisfying the near-zero false-reassurance criterion highlighted in the Introduction.  Its lightweight architecture and transparent training protocol make integration into ICU decision-support pipelines feasible.

\paragraph{Limitations and future work.}
This study is retrospective and single-centre; external validation across multiple institutions and EEG systems is required.  Future efforts will explore multimodal fusion with clinical variables, probabilistic calibration for uncertainty-aware predictions, and continual learning to accommodate evolving care practices.

\paragraph{Conclusion.}
When patient-level leakage is rigorously prevented, deep neural models can extract actionable prognostic information from raw EEG.  Our two-stage framework offers a reliable, computation-efficient foundation for early outcome prediction and paves the way for clinically deployable EEG analytics in critical care.

\end{document}